\documentclass[lettersize,journal]{IEEEtran}
\usepackage{amsmath,amsfonts}
\usepackage{algorithmic}
\usepackage{algorithm}
\usepackage{array}
\usepackage[caption=false,font=normalsize,labelfont=sf,textfont=sf]{subfig}
\usepackage{textcomp}
\usepackage{stfloats}
\usepackage{url}
\usepackage{verbatim}
\usepackage{graphicx}
\usepackage{cite}
\usepackage{booktabs}
\usepackage{xcolor}
\usepackage{tabularx}
\usepackage{ragged2e}
\usepackage{multirow} 
\usepackage{threeparttable}
\usepackage{microtype}

\newcolumntype{Y}{>{\RaggedRight\arraybackslash}X}

\newif\ifcomment
\commenttrue

\newcommand{\shih}[1]{\ifcomment{\color{red}}\else\fi}

\begin{document}

\title{Temporal Hyperbolic Graph Representation Learning for Scale-Free Internet Routing and Delay Prediction}

\author{\shih{is not complete yet}
    Yi-Ling Kuo\textsuperscript{1}, 
    Hao-Yu Tien\textsuperscript{1}, 
    Shih-Yu Tsai\textsuperscript{1,*} 
    
    \textsuperscript{1} Department of Information Management and Finance, National Yang Ming Chiao Tung University, Hsinchu 30010, Taiwan\\[4pt]
    
    Emails: 0805linda2.0@gmail.com (Y.L. Kuo), 
    stu710319@gmail.com (H.Y. Tien), \\
    \hspace*{1.5em}shih-yu.tsai@nycu.edu.tw (S.Y. Tsai)\\
    *Corresponding author
}



\maketitle

\begin{abstract}\shih{fig table  position change}\shih{compact the abstact}\shih{before submission, put all content into grammarly for polishing}
Predicting Internet round-trip time (RTT) is critical for routing optimization, quality-of-service (QoS) provisioning, and traffic engineering in large-scale networks, yet it remains challenging because RTT depends on a long-term history, follows complex routing dynamics, and exhibits heavy-tailed latency distributions. While Temporal Graph Neural Networks (TGNNs)~\cite{longa2023graph, you2022roland, rossi2020temporal, pareja2020evolvegcn, sankar2020dysat, yang2022hyperbolic, nickel2017poincare, le2024toward, Xu2020Inductive} have shown promise for modeling evolving network topologies, most existing approaches operate in Euclidean space, which is not well-suited for representing the hierarchical and scale-free structure of Internet routing graphs. In contrast, hyperbolic geometry provides a natural embedding space for such topologies with low distortion in low dimensions~\cite{sarkar2011low}.

In this work, we propose \textbf{HERMIT}(Hyperbolic Edge-aware RTT modeling via Integrated Topology), a hybrid framework that integrates a hyperbolic manifold-preserving temporal GNN with a tree-based Random Forest regressor for joint link prediction and RTT prediction. Building on HMPTGN~\cite{le2024toward} as the backbone, we further enhance its temporal graph encoder by introducing RTT-aware edge features and a learnable edge encoder, allowing more precise modeling of the link states and evolving routing behavior. We combine the resulting hyperbolic node representations with historical RTT statistics in a Random Forest regressor to model non-linear and heavy-tailed latency patterns.

We evaluated HERMIT on a large-scale real Internet dataset spanning 10 years from 2015 to 2024. Our experiments show that HERMIT consistently outperforms a strong Random Forest baseline~\cite{stepanov2025round} based solely on historical RTT statistics, achieving a 6\% improvement in RMSE and substantially mitigating large errors on heavy-tailed samples. Moreover, HERMIT, with its hyperbolic temporal representations enhanced by RTT-aware edge features, achieves better link prediction than prior hyperbolic TGNN models, including HMPTGN~\cite{le2024toward} and HTGN~\cite{yang2022hyperbolic}, highlighting the effectiveness of incorporating RTT-aware edge modeling. Our results suggest that combining hyperbolic temporal graph representations with tree-based regression provides a robust and scalable solution for RTT prediction in the real-world Internet topologies.
\end{abstract}

\shih{for shih: ack yimin rezaul jie

possible committee member:
Yen-Chang Huang:
\url{https://sites.google.com/view/ychuang/home}
\url{https://www.ee.ntu.edu.tw/profile1.php?id=1090744}
\url{https://sites.google.com/view/eli-chien/home}
\url{https://docs.google.com/presentation/d/10d2Whr_1I3gcX5ToBAgw_s6blzYcru_N/edit?slide=id.p61#slide=id.p61}
\url{https://drive.google.com/file/d/12qN9QmbdhcrNpmq7sCehbnkQzJBGD-sP/view}}
\section{Introduction}

Round-trip time (RTT) is a fundamental metric for characterizing network performance, measuring the time required for a packet to travel from a source node to a destination and back. RTT directly impacts multiple aspects of network operation: from a user experience perspective, excessive RTT leads to slow page loads, video conferencing lag, and online gaming delays, severely degrading the usability of interactive applications; from a network engineering perspective, ISPs leverage RTT data to identify high-utilization paths during peak hours, informing decisions on bandwidth upgrades, peering adjustments, or traffic engineering strategies to redistribute load and optimize overall network performance. Accurate RTT prediction is therefore critical for Quality of Service (QoS) guarantees—for instance, CDNs require predictive low-latency path selection, and cloud services need dynamic resource allocation—yet existing methods struggle to maintain stable prediction accuracy under both normal and abnormal RTT conditions.

 We are going to design a model that can jointly address link prediction and RTT prediction. However, RTT modeling and prediction face fundamental challenges. Prior studies have shown that RTT measurements in the Internet exhibit high variability and heavy-tailed distributions\cite{fontugne2015empirical, salimi2022analysis}. Our dataset aligns with these observations: RTT values span sub-millisecond low-latency links to extreme cases approaching 10 seconds. This extreme distributional characteristic introduces two challenges: First, heavy tails cause traditional mean squared error (MSE)-based regression models to be dominated by outliers, degrading prediction accuracy in the common RTT range. Second, RTT is influenced by routing dynamics, congestion states, and topology evolution, exhibiting highly non-linear and time-varying behavior that is difficult to capture with conventional regression models and static-distance embedding approaches.
To address these challenges, early RTT prediction methods primarily relied on geometric coordinate systems, such as the Vivaldi algorithm~\cite{dabek2004vivaldi}, which embeds nodes into coordinate spaces to approximate latencies, or latent factorization techniques like DMFSGD~\cite{liao2012dmfsgd}, which captures implicit node-pair relationships through matrix decomposition. These approaches can capture coarse spatial relationships, but they are designed for static or quasi-static environments and struggle to handle dynamic routing changes, transient congestion, and highly non-linear latency dynamics.

Recently, machine learning-based approaches have emerged, including a sequence-based model, which is an LSTM architecture that treats dynamic RTT as a time series~\cite{hagos2019deep}, and a tree-based ensemble model, Random Forest~\cite{stepanov2025round}. These methods improve regression accuracy by capturing temporal dependencies as well as non-linear relationships and interactions among handcrafted network features (e.g., historical RTT statistics, path characteristics, and traffic-related attributes). However, they typically treat RTT prediction as a tabular regression problem (i.e., predicting a numerical value given a set of attributes/features), neglecting the underlying network topology; the RTT between the sender nodes and the receiver nodes is actually affected by structural factors such as routing paths, congestion status of intermediate nodes, and topology evolution. This disregard for graph structure limits the models' ability to generalize to the scenarios of routing changes and emerging connections. In order to effectively utilize this topology information, Jaeger et al.~\cite{jaeger2022modeling} modeled Transmission Control Protocol (TCP) behavior using Graph Neural Networks (GNNs), encoding the network topology as a graph and predicting throughput and RTT simultaneously. Their experimental results show that GNN can demonstrate better generalization capability across different network topologies and TCP congestion control algorithms, and can effectively learn the relationships between TCP performance metrics such as RTT and throughput. However, the above methods mostly treat the network as a static graph and do not explicitly capture how network states evolve over time.

Temporal Graph Neural Networks (TGNNs)~\cite{longa2023graph, you2022roland, rossi2020temporal, pareja2020evolvegcn, sankar2020dysat, yang2022hyperbolic, nickel2017poincare, le2024toward, Xu2020Inductive} have recently attracted increasing attention because they are capable of modeling the temporal dynamics of network performance with better quality. However, we observe that most existing temporal graph neural networks used for network modeling are still designed and implemented in Euclidean space~\cite{longa2023graph, you2022roland, rossi2020temporal, pareja2020evolvegcn, sankar2020dysat}, whereas architectures specifically designed for hyperbolic geometry, which is better suited to represent the topology of the Internet~\cite{krioukov2010hyperbolic, sarkar2011low}. The Internet topology exhibits both scale-free degree distributions and hierarchical structure. When embedding such graphs in Euclidean space, it often requires high embedding dimensionality or incurs geometric distortion to accommodate the large number of lower-layer nodes. The volume of Euclidean space grows only polynomially with the radius, making it difficult to preserve distances and adjacency relations of large-scale hierarchical networks in low dimensions.
With exponential volume growth of hyperbolic geometry, low-dimensional embeddings can naturally accommodate tree-like and scale-free structures whose number of nodes grows exponentially with depth~\cite{krioukov2010hyperbolic, yang2022hyperbolic}. Prior work~\cite{nickel2017poincare, chami2019hyperbolic, liu2019hyperbolic} has shown that embedding hierarchical graphs in hyperbolic space can significantly reduce distortion and preserve both hierarchical relations and long-range relative positions between nodes. Thus, designing temporal graph neural networks in hyperbolic space is more suitable for hierarchical and scale-free network data, such as Internet routing graphs, providing more stable and expressive representations than traditional Euclidean TGNNs in the same dimension.

With these observations, in order to get better accuracy and robustness performance on link connection prediction and the corresponding RTT values estimate, we proposed a novel hybrid architecture, called \textbf{HERMIT}, that combines a manifold-preserving hyperbolic temporal GNN with edge-level feature and a tree-based regression model.
Building on HMPTGN~\cite{le2024toward}, we introduce two key enhancements to the graph encoding module. First, we incorporate explicit edge features that fuse RTT statistics to characterize fine-grained link states. Second, we design a learnable edge encoder that adaptively assigns weights to different links during message passing, thereby emphasizing edges that are more critical, stable, and frequently observed in the topology. With these extensions, our model learns node embeddings that more finely reflect link states and path behavior, and further improves link prediction performance over HMPTGN~\cite{le2024toward} and HTGN~\cite{yang2022hyperbolic}. For the RTT prediction task, we combine information from the learned graph structure and historical RTT statistics. Specifically, the  hyperbolic embeddings generated by our hyperbolic temporal encoder are treated as topological features. They are directly concatenated with historical RTT statistics and then fed into a Random Forest regressor to produce RTT estimates. Compared to regression models trained only on historical RTT statistics~\cite{stepanov2025round}, this hybrid feature design enables the regressor to exploit topology-aware information from hyperbolic graph embeddings, yielding better overall performance in RTT prediction. The hyperbolic temporal encoder captures macro-level network organizations, while the Random Forest exploits micro-level feature interactions, resulting in more accurate and robust RTT predictions than either approach alone.


Our main contributions are summarized as follows:
\begin{itemize}
    \item We design a reproducible data preprocessing pipeline that transforms raw CAIDA Ark traceroute logs~\cite{caida_ipv4_prefix_probing} into a sequence of daily weighted graphs with edge-level RTT statistics. The pipeline addresses data inconsistencies and incomplete observations commonly observed in Internet traceroute measurements~\cite{fomenkov2011datamanagement}.
    \item We propose \textbf{HERMIT}, a sate-of the-art RTT prediction framework that combines a hyperbolic manifold-preserving temporal GNN with a tree-based regressor. We demonstrate that it can handle massive Internet data by  evaluating its RTT prediction performace on a large-scale, real-world dataset spanning ten years (2015 - 2024) of U.S. Internet measurements from the CAIDA IPv4 Ark project~\cite{caida_ipv4_prefix_probing}.
    \item For RTT prediction, our model HERMIT combines hyperbolic topology-aware embeddings with historical RTT information and achieves significant improvements in RMSE by \textbf{6\%} compared to a Random Forest model~\cite{stepanov2025round} that uses only historical RTT statistics. Specifically, it is by \textbf{6.4\%} on existing links and by \textbf{2.3\%} on new appearing links. 
    \item For link prediction, our model (HERMIT) that incorporates explicit edge features and a learnable edge encoder consistently outperforms prior hyperbolic GNN baselines including HMPTGN~\cite{le2024toward} and HTGN~\cite{yang2022hyperbolic}, on both existing and new links, achieving AUC and AP scores above 0.99 and demonstrating the effectiveness of the learned hyperbolic temporal representations.
    \end{itemize}
Overall, our results demonstrate that combining hyperbolic temporal graph representations with Random Forest regression provides a robust and scalable solution for RTT prediction under real-world, heavy-tailed network latency conditions.

\shih{write it in a compact way}The paper is structured as follows. Section~\ref{notation} introduces the problem and introduces the dynamic graph setting and learning tasks. Section~\ref{relatedwork} reviews related work. Section~\ref{dataset} describes the dataset and preprocessing.
Section~\ref{methodology} presents the proposed HERMIT framework. Section~\ref{experiment} details the experimental setup, including dataset splits, baselines, and evaluation metrics. Section~\ref{results} reports the empirical results on link prediction and RTT prediction. Section~\ref{discussion} discusses the findings. Section~\ref{conclusion} concludes the paper.

\section{Notation, Problem Formulation, and Preliminaries}
\label{notation}

In this section, we formally define the dynamic graph setting that is considered and introduce the learning tasks of interest. Specifically, we first describe the representation of the Internet routing network as a sequence of time-ordered graph snapshots with associated edge attributes. We then formulate three core tasks: (1) link prediction, which estimates the likelihood of future connections; (2) new link prediction, which focuses on forecasting previously unseen routing relationships; and (3) RTT prediction, which predicts the future continuous RTT values of network links. These tasks together enable comprehensive modeling of both structural dynamics and latency evolution in time-varying Internet graphs.

\subsection{Notation}
Scalars (e.g., the time index $t$, an RTT value $y$) are denoted by regular letters, and vectors by boldface lowercase letters (e.g., $\mathbf{z}$). A hat indicates a predicted or estimated quantity, for example $\hat{y}$ denotes the predicted RTT value corresponding to the true RTT value $y$.

For a node $u$ at time $t$, we write $\mathbf{z}_u(t)$ for its hyperbolic node embedding at that time, that lies in a $d$-dimensional hyperbolic space (e.g., the Poincar\'e ball model), and the detailed hyperbolic configuration setting will be introduced in Section~\ref{experiment}.We also use $\mathbf{f}_u$ and $\mathbf{f}_v$ to denote the historical RTT feature vectors of nodes $u$ and $v$, respectively.

\subsection{Definition of Dynamic Graph}

A dynamic graph is a graph whose structure and attributes evolve over time. In our work, we adopt a \emph{discrete-time snapshot} formulation, where the dynamic graph is represented as a sequence of graph snapshots $\{G_1, G_2, \ldots, G_T\}$. Each snapshot at time $t$
\[
G_t = (V_t, E_t, X_t)
\]
consists of a set of nodes $V_t$, a set of directed edges $E_t \subseteq V_t \times V_t$, and associated edge features $X_t$. In particular, $X_t$ contains link-level measurements (e.g., RTT and other metrics) observed at time $t$ for every edge in $E_t$, so that each snapshot represents the Internet routing topology together with RTT and related link-level metrics collected during a given time window.

In our setting, both the edge set $E_t$ and the edge features $X_t$ may change over time as routes appear, disappear, or vary in performance, and the node set $V_t$ can also differ across snapshots.
For any two distinct nodes $u, v \in V_t$ with $u \neq v$, a directed edge $(u,v) \in E_t$ represents a routing connection from $u$ to $v$ at time $t$. The feature vector for each edge $(u,v) \in E_t$ is denoted by
\[
\mathbf{x}_{uv,t} \in \mathbb{R}^{3},
\]
which includes the log-transformed mean RTT, log-transformed RTT standard deviation, and the link weight.

\subsection{Link prediction}

The link prediction task aims to infer which node pairs are likely to be connected in the next snapshot based on historical graph structure and edge attributes. Formally, given a sequence of past snapshots $\{G_1, G_2, \ldots, G_t\}$, the goal is to estimate, for each candidate node pair $(u,v)$ in $V_t \times V_t$ where $u \neq v$, the probability that an edge $e_{uv}$ exists in the future snapshot $G_{t+1}$, i.e.,
\[
\hat{p}_{uv,t+1} = \Pr\big[(u,v) \in E_{t+1} \,\big|\, G_1, \ldots, G_t\big].
\]
This formulation evaluates the model on all edges that appear in $G_{t+1}$, including both edges that persist from earlier snapshots and newly formed connections, providing a measure of the model's overall predictive capability across the entire network topology.

\subsection{New Link Prediction}

New link prediction focuses on edges that appear for the first time in $G_{t+1}$. 
We define the set of newly formed edges as $E_{t+1}^{\text{new}} = E_{t+1} \setminus \cup_{i=1}^{t} E_i$. For each $(u,v) \in E_{t+1}^{\text{new}}$, we estimate:
\[
\hat{p}_{uv,t+1}^{\text{new}} = \Pr\big[(u,v) \in E_{t+1} \,\big|\, (u,v) \notin E_t, G_1, \ldots, G_t\big].
\]
This formulation measures the model's generalization ability to predict emerging connections, rather than memorize recurring ones.

\subsection{RTT prediction}
RTT prediction aims to predict a real-valued RTT for a given network edge. Given past snapshots $\{G_1, G_2, \ldots, G_t\}$ and a target node pair $(u,v)$, the goal is to estimate the RTT value in the next snapshot $G_{t+1}$, denoted by
\[
\hat{y}_{uv,t+1},
\]
capturing the conditional expectation of RTT given the historical graph and measurements. This task models the temporal dynamics of latency on both existing and newly appearing routing links.

\section{Related Work}
\label{relatedwork}
This section reviews the evolution of two distinct research domains, dynamic graph representation learning and network latency prediction, that form the foundation of our work. 
\subsection{Dynamic Graph Representation Learning}
Dynamic graph representation learning~\cite{kazemi2020representation, skarding2021foundations} focuses on modeling networks that evolve over time by learning embeddings for nodes or edges. Given a sequence of graph snapshots, the objective is to capture both the structural patterns and their temporal changes, enabling the learned representations to be used in downstream tasks such as link prediction and RTT prediction.
\shih{author year writting!}
Early approaches treated dynamic graphs as a sequence of independent static snapshots and learned Euclidean embeddings for each snapshot separately. In 2014, Perozzi et al. proposed DeepWalk~\cite{perozzi2014deepwalk} ,which introduced random-walk-based representation learning, which was later extended in 2017 by GraphSAGE ~\cite{hamilton2017inductive} proposed by Hamilton et al., to neighborhood aggregation methods. These methods were applied to each time-stamped graph to learn node embeddings for temporal link prediction. While these snapshot-based Euclidean models are simple and scalable, they often overlook temporal dependencies and smooth temporal transitions.

As temporal modeling became more important in the early 2020s, prior work incorporated explicit temporal modeling into Euclidean embedding frameworks~\cite{longa2023graph, rossi2020temporal}. In particular, as surveyed in~\cite{longa2023graph}, early approaches leveraged dynamic matrix factorization and recurrent architectures (e.g., RNNs and GRUs) to enforce temporal consistency in node representations. Building on these ideas, representative dynamic GNNs such as EvolveGCN~\cite{pareja2020evolvegcn}, TGAT~\cite{Xu2020InductiveRL}, and Roland~\cite{you2022roland} combine graph neural networks with recurrent or attention-based mechanisms to model structural and temporal dependencies.
On the other hand, event-based methods model dynamic graphs as continuous-time interaction streams. In 2020, Rossi et al. proposed Temporal Graph Networks (TGN)~\cite{rossi2020temporal} update node embeddings upon each observed event, providing finer temporal resolution and strong performance on dynamic link prediction. However, these methods are often computationally expensive for long-term, large-scale data, making snapshot-based approaches more suitable for our setting.

Despite these advances, Euclidean embeddings struggle to represent networks with strong hierarchical or tree-like structures, such as Internet routing topologies, often leading to distorted distances or requiring high-dimensional representations. This limitation has motivated the adoption of hyperbolic geometry, which can embed such structures more effectively with lower dimensionality while preserving hierarchical relationships~\cite{yang2022hyperbolic, nickel2017poincare}.

Building on these insights, Yang et al. introduced Hyperbolic Temporal Graph Networks (HTGN)~\cite{yang2021discrete} in 2021, one of the first frameworks to incorporate hyperbolic geometry into temporal graph modeling. HTGN maps temporal graphs into a hyperbolic latent space and combines hyperbolic graph neural networks with hyperbolic recurrent units to capture both temporal dynamics and underlying hierarchical structures, demonstrating strong performance on temporal link prediction tasks.

However, HTGN still relies on operations in the tangent (Euclidean) space, requiring frequent mappings between Euclidean and hyperbolic spaces. This back-and-forth transformation can introduce geometric distortion and limit the model’s ability to fully exploit the properties of hyperbolic geometry~\cite{yang2021discrete, le2024toward}.

To address this limitation, Le et al. proposed Hyperbolic Manifold-Preserving Temporal Graph Networks (HMPTGN)~\cite{le2024toward} in 2024, which perform message passing directly on the hyperbolic manifold. By avoiding intermediate projections to tangent space, HMPTGN reduces geometric distortion and provides a more faithful representation of spatio-temporal dependencies in hierarchical networks.

Building upon this line of work, our approach further extends hyperbolic temporal graph modeling toward Internet routing topologies and RTT prediction tasks.

\subsection{Network Latency and RTT Prediction}
Classical TCP packet measurement implementation methods can be dated back to the 1980s, Jacobson estimated RTT using the Jacobson-style exponentially weighted moving average (EWMA)~\cite{jacobson1988congestion}. In the early 2000s, RTT prediction methods primarily relied on Euclidean coordinate systems~\cite{dabek2004vivaldi} or latent factorization techniques~\cite{liao2012dmfsgd}, which  capture coarse spatial relationships (i.e., approximate distances between network nodes). However, these approaches struggle to model the non-linear and transient variations that characterize modern internet latency dynamics.

Motivated by these limitations, data-driven deep learning methods were introduced to capture temporal patterns in RTT sequences. In 2019, Hagos et al. \cite{hagos2019deep} proposed an LSTM-based dynamic passive RTT prediction model for TCP and showed that they can closely track end-to-end RTT in real time, achieving low prediction error in their experimental TCP traffic scenarios. 

Recent work on RTT prediction using classical machine learning models highlights tree-based ensembles as strong baseline models. In particular, in 2025, Stepanov et al.~\cite{stepanov2025round} empirically demonstrate that Random Forests outperform regularized linear models (e.g., ElasticNet~\cite{zou2005regularization}), Jacobson’s EWMA-based RTT estimator, and recurrent neural networks (RNN, LSTM, GRU) in terms of MSE, MAE, and MAPE across diverse traffic scenarios, including heavy-tailed RTT distributions.

However, these sequence-based models remain constrained due to their end-to-end formulation. Specifically, treating RTT prediction as an isolated temporal regression task means these models ignore how the underlying network topology and routing paths influence latency. 
Existing approaches partially address these aspects but remain incomplete. Hyperbolic temporal GNNs (e.g., HTGN~\cite{yang2021discrete} and HMPTGN~\cite{le2024toward}) capture hierarchical structure and dynamic connectivity, yet do not explicitly model fine-grained RTT variations. On the other hand, state-of-the-art RTT predictors (e.g., sequence models~\cite{hagos2019deep} and tree-based ensembles~\cite{stepanov2025round}) focus on temporal patterns while overlooking the underlying network geometry.

Therefore, it motivates us to propose the HERMIT framework, which bridges hyperbolic graph learning and latency prediction. By jointly modeling topological evolution and temporal delay dynamics on the hyperbolic manifold, HERMIT leverages the geometric hierarchy of Internet networks to produce more accurate and reliable RTT estimates.

\shih{move it so that it is after method sec ?}

\section{Dataset}
\label{dataset}
In this section, we show how we create our dataset. The utilized dataset originates from the CAIDA IPv4 Ark Project~\cite{caida_ipv4_prefix_probing}, specifically from the probe data operated by Team-1.
The temporal scope of the dataset is from 2015 to 2024, focusing primarily on probing activities within the United States.

The raw dataset comprises active traceroute measurements stored in the binary .warts format. In order to analyze, we parse the traces and convert them into a JSON file to facilitate feature extraction and temporal graph construction. The parsed dataset contains various attributes for each trace path from a source probe (src) to a destination address (dst). In fact, we only care about key attributes for our analysis, as shown in Table~\ref{tab:data_schema}.
The remaining extracted attributes can be classified into two categories: metadata and topology data. Metadata includes \texttt{start} and \texttt{stop\_reason}. The \texttt{start} field records the precise timestamp of when each traceroute probe is initiated, enabling temporal alignment across measurements. The \texttt{stop\_reason} (e.g., COMPLETED, LOOP, UNREACH) indicates the termination status of each trace. 
Topology data includes \texttt{hop\_count} and the \texttt{hops} array. The \texttt{hop\_count} records the total number of hops traversed in the path. The \texttt{hops} array contains an ordered sequence of intermediate routers traversed during the traceroute process. Each element in the \texttt{hops} array represents a hop, and in each hop, it has the visited router's IP address (\texttt{addr}) and the corresponding round-trip time (\texttt{rtt}). Thereby, it provides the essential information for constructing nodes, edges, and latency measurements in the network graph. Importantly, this hop sequence is inherently ordered, reflecting the step-by-step propagation of packets along the network path from source to destination.

\begin{table}[h!]
    \centering
    \caption{Description of Key Fields in the CAIDA Ark Dataset}
    \label{tab:data_schema}
    
    \begin{tabular}{l l p{4.5cm}} 
        \toprule
        \textbf{Field Name} & \textbf{Data Type} & \textbf{Description} \\
        \midrule
        
        \texttt{src} / \texttt{dst} & String & 
        The IPv4 addresses of the source probe and the destination target. \\
        \addlinespace[1ex] 
        
        \texttt{start} & Object & 
        Contains the measurement timestamp (\texttt{sec}), used to construct temporal graph snapshots. \\
        \addlinespace[1ex]
        
        \texttt{stop\_reason} & String & 
        Indicates the termination status of the trace (e.g., \texttt{COMPLETED}, \texttt{LOOP}, \texttt{UNREACH}). \\
        \addlinespace[1ex]
        
        \texttt{hop\_count} & Integer & 
        The total number of hops traversed in the path. \\
        \addlinespace[1ex]
        
        \texttt{hops} & Array & 
        A sequential list of intermediate routers. Each element contains: \newline
        \textbullet~ \texttt{addr}: Router IP address (Node). \newline
        \textbullet~ \texttt{rtt}: Round-Trip Time. \\
        
        \bottomrule
    \end{tabular}
\end{table}

Next, we clean and organize the raw dataset into temporal graphs for model training and evaluation. To handle the large volume of raw data, an initial preprocessing step was applied to remove incomplete traces. We only retain measurement of probes  whose \texttt{stop\_reason} is labeled as "completed", ensuring that the inferred network topology data is based on valid paths where the traceroute reached its destination.
To mitigate the highly skewed degree distribution characteristic of network topologies, the analysis was restricted to the most connected nodes. Specifically, the top $5\%$ of nodes ranked by degree were selected, which collectively accounted for approximately $90\%$ of the overall connection volume. This pruning step preserves the core backbone structure while discarding low-active nodes that contribute little to the aggregate dynamics.

Even after this pruning reduction, our temporal dataset led to excessive computational overhead and GPU  memory exhaustion during model training (we use a NVIDIA GeForce RTX 4090 GPU with 24 GB of memory) In order to maintain computational feasibility and temporal variability of the dataset at the same time, we adopt a rolling weekly sampling strategy. Instead of using all the daily data, we sample three days per week. It contains two weekdays and one weekend day for each week. The specific sampling days were alternated based on the week number. That is Monday, Wednesday, and Saturday in odd-numbered weeks; Tuesday, Friday, and Sunday in even-numbered weeks. This provides a balanced and reasonable subset that covers both typical weekday behavior and weekend traffic patterns. This scheme reduces the original dataset, which consists of 3,415 temporal snapshots to 1,456 snapshots (a 42.6\% reduction), while still maintaining both consistent weekly coverage and representative diurnal traffic patterns.

After preprocessing and aggregation, we get our final input data structured into six key fields as detailed in Table~\ref{tab:dataset_fields}. The data encapsulates traceroute measurements collected within the United States from 2015 to 2024, capturing the daily dynamics of network-level interactions, as detailed in Table~\ref{tab:dataset_fields}.

\begin{table}[h!]
    \centering
    \caption{Description of the Preprocessed Dataset Fields}
    \label{tab:dataset_fields}
    \begin{tabular}{l l p{5cm}} 
        \toprule
        \textbf{Field} & \textbf{Type} & \textbf{Description} \\
        \midrule
        \texttt{source} & Integer & Anonymized ID representing the starting node of a network hop. \\
        \texttt{target} & Integer & Anonymized ID representing the destination node of the hop. \\
        \texttt{time}   & Integer & Discrete timestamp index in the sequence of sampled snapshots. \\
        \texttt{weight} & Integer & Aggregated frequency count of the link observed on that given day. \\
        \texttt{avg\_rtt}& Float   & Average Round-Trip Time (ms) of the link for the day, indicating latency. \\
        \texttt{std\_rtt}& Float   & Standard deviation of RTT (ms), capturing the stability of the link. \\
        \bottomrule
    \end{tabular}
\end{table}

We construct a sequence of temporal graphs $\{G_t\}_{t=1}^T$, where each daily snapshot $G_t = (V_t, E_t, X_t)$ represents the Internet topology and latency statistics on day $t$. For each day $t$, the node set $V_t$ consists of all \texttt{source} and \texttt{target} IDs observed on that day, and the edge set $E_t$ contains a directed edge $(u,v)$ if at least one measurement with \texttt{source} $=u$, \texttt{target} $=v$, and \texttt{time} $=t$ is observed. The edge attributes $X_t$ include the aggregated frequency \texttt{weight}, the daily average RTT \texttt{avg\_rtt}, and the RTT standard deviation \texttt{std\_rtt}. The model takes as input a sequence of daily snapshots $\{G_1 \dots, G_T\}$, where each $G_t$ encodes both the routing topology and historical RTT statistics for day $t$.

\begin{figure*}[!t]
    \vspace{-6pt}
    \centering
    \includegraphics[width=1\linewidth]{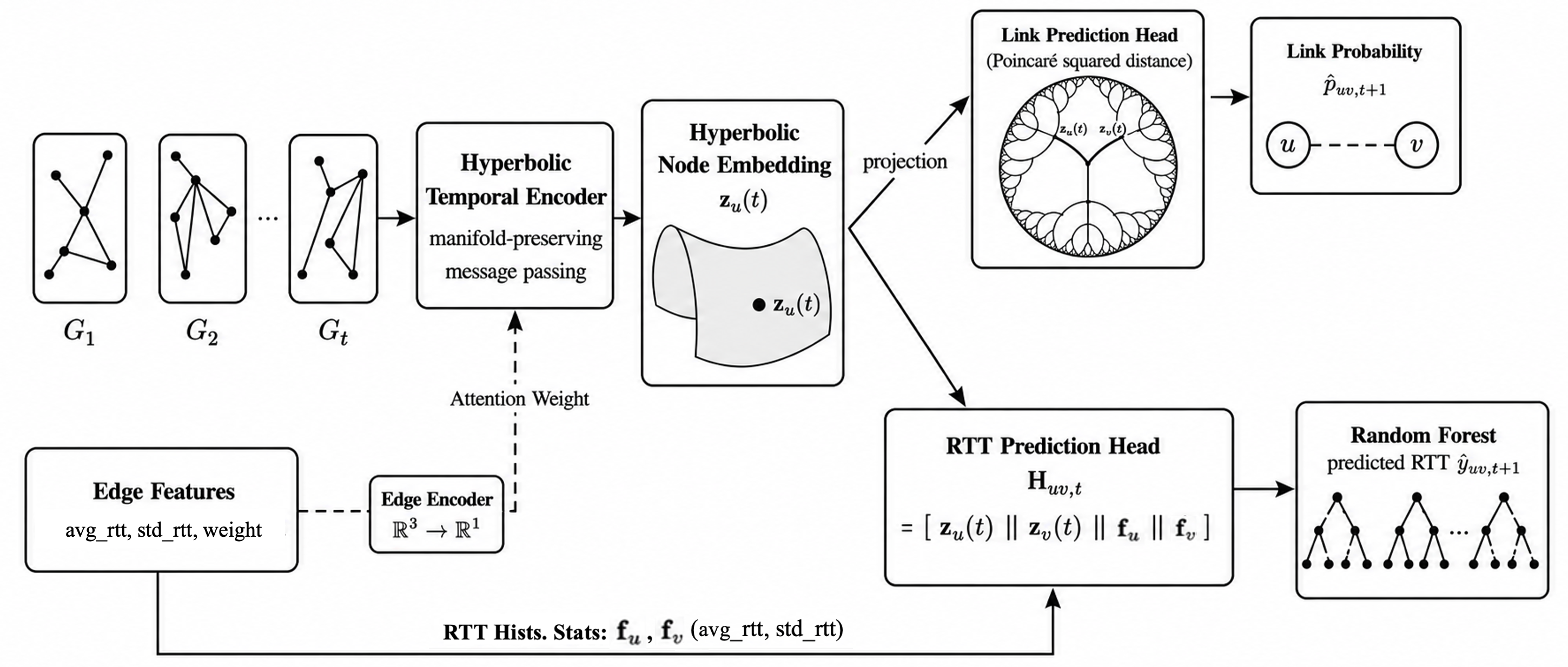}
    \caption{Overview of the proposed HERMIT architecture. Based on HMPTGN, our hyperbolic temporal encoder incorporates edge-level features through a learnable edge encoder. The resulting hyperbolic embeddings are jointly utilized for link prediction and RTT prediction via Random Forest regression.}
    \label{fig:architecture}
\end{figure*}

\section{Methodology}
\label{methodology}
We propose HERMIT, a hybrid architecture that extends HMPTGN~\cite{le2024toward} with a learnable edge encoder and Random Forest for joint link and RTT prediction. The key design principle is to preserve the hierarchical structure of Internet topology in hyperbolic space while incorporating explicit edge features, and to leverage the robustness of Random Forest for handling heavy-tailed RTT distributions. As illustrated in Fig.~\ref{fig:architecture}, HERMIT performs this joint prediction task on sequences of daily network snapshots.
\shih{why project them into tangent space not directly do it? because consider all of them into tangent space (emb + rtt value) or can we do rtt value into hyperbolic space? I mean take rtt into several diff classes 0-1 1-2 2-4 4-8 8-16 etc to come up a random forest for rtt value?}

\label{design rationale}
The HMPTGN is selected as the backbone architecture due to its superior geometric stability and enhanced feature representation capabilities compared to the baseline HTGN, which are critical for accurately modeling the highly dynamic and hierarchical Internet topology. The primary distinction between these two models lies in HMPTGN's manifold-preserving design, which utilizes native hyperbolic operations such as Möbius addition and Poincaré linear transformations to minimize geometric distortion and ensure critical structural information is accurately preserved. Furthermore, HMPTGN enhances stability by applying orthogonal weight matrices, which constrain node embeddings within the Poincaré ball during their mathematical operations, effectively mitigating numerical instability observed under aggressive optimization settings.
 Conversely, the baseline HTGN model relies on tangent space approximation, projecting embeddings onto a local Euclidean tangent plane, an approach that introduces significant geometric distortion and allows the resulting error to accumulate linearly across time steps, leading to numerical instability. 
\label{edge}
\begin{table}[h!]
    \centering
    \caption{Node and Edge Feature Engineering and Preprocessing Pipeline}
    \label{tab:feature_engineering}
    \renewcommand{\arraystretch}{1.1}

    \begin{tabularx}{\linewidth}{@{}l c Y Y@{}}
        \toprule
        \textbf{Feature} & \textbf{Dim} & \textbf{Content} & \textbf{Preprocessing Pipeline} \\
\midrule
\textbf{Node Feature} & 128 
& Learnable node embedding
& Jointly learned during training \\
\midrule
\textbf{Edge Feature} & 3 
& Log-normalized RTT, log-normalized RTT standard deviation, constant weight
& Log transform $\rightarrow$ global min--max normalization (scaled to [0, 1]) \\
\midrule
        \textbf{Edge Encoder} & 3 $\rightarrow$ 1 
        & Linear transformation + sigmoid
        & Xavier initialization $\rightarrow$ jointly optimized during training \\
    \bottomrule
    \end{tabularx}
\end{table}

Following the original HMPTGN architecture, node features are initialized as learnable embeddings of dimension 128 that are jointly optimized with the model parameters to adaptively capture the  topological structure without manual feature engineering. To more comprehensively represent the network dynamics, this node-centric approach is extended by incorporating edge features, which explicitly encode the link quality by incorporating the log-normalized metrics of RTT mean and standard deviation, along with the link weight. (Note that these edge features are scaled into $[0,1]$ via global Min-Max normalization to preserve the absolute scale of RTT across all temporal snapshots.) To effectively integrate these three-dimensional edge features, we employ a learnable linear transformation from $\mathbb{R}^3$ to $ \mathbb{R}^1$ followed by a sigmoid activation to enable learnable attention weights.
This edge encoder acts as an edge-level attention mechanism, compressing the three-dimensional edge attributes into a scalar weight in $[0, 1]$, which is used to adjust the message passing in the HERMIT layers. The overall feature engineering and preprocessing pipeline for nodes and edges is summarized in Table~\ref{tab:feature_engineering}.
\begin{figure}
    \centering
    \includegraphics[width=1\linewidth]{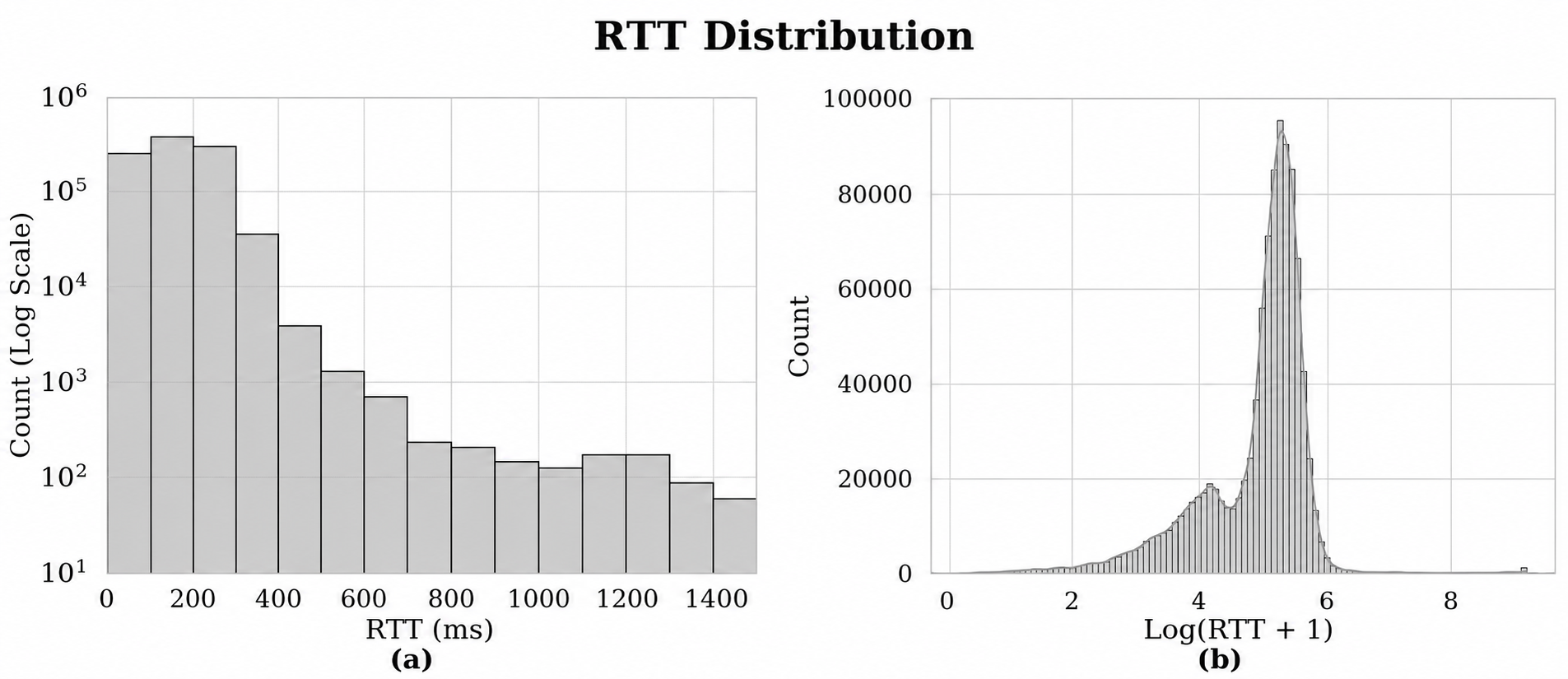}
    \caption{(a) RTT distributions on a linear scale show heavy-tailed characteristics with most values below 200ms but outliers extending beyond 1000ms. (b) The Gaussian like log-transformed RTT distribution exhibits more symmetric and concentrated characteristics suitable for regression.}
    \label{fig:rtt_distribution}
\end{figure}

As illustrated in Fig.~\ref{fig:rtt_distribution}(a), the RTT values in our dataset exhibit a heavy-tailed distribution, with most falling below 200 ms and only a small fraction exceeding 1000ms. Since a standard MSE objective is highly sensitive to these extreme outliers, it often degrades prediction accuracy for typical short-RTT connections. As in Fontugne et al.~\cite{fontugne2015empirical}, who model RTTs in the log-domain to handle heavy-tailed behavior, we apply a logarithmic transformation $\log(\mathrm{RTT} + 1)$ followed by min–max normalization to compress outliers and obtain a more symmetric distribution, as illustrated in Fig.~\ref{fig:rtt_distribution}(b). Using $\log(\mathrm{RTT} + 1)$ ensures that the transformed values are positive and avoids extreme negatives from near-zero RTTs, which stabilizes normalization and model training. Our model is trained to predict the RTT values in this normalized logarithmic space. At evaluation time, we perform an inverse Min-Max transformation and exponentiate the predictions to report RMSE and MAE on the original millisecond scale.

\label{proposed HERMIT architecture}
As Fig.~\ref{fig:architecture} shows, our framework consists of a hyperbolic temporal encoder and an RTT prediction module that regresses continuous RTT values from hyperbolic node embeddings and historical latency statistics, in addition to the link probability used in the original decoder. 
The encoder first maps graph snapshots into the Poincaré ball, then performs edge-aware hyperbolic message passing to aggregate neighbor features, followed by a GRU to capture temporal dynamics. At each timestamp $t$, it produces hyperbolic node embeddings $\mathbf z_u(t)$ for all nodes $u$ that preserve the hierarchical structure of Internet topology.

\begin{figure*}[h!]
    \vspace{-6pt}
    \centering
    \includegraphics[width=1\linewidth]{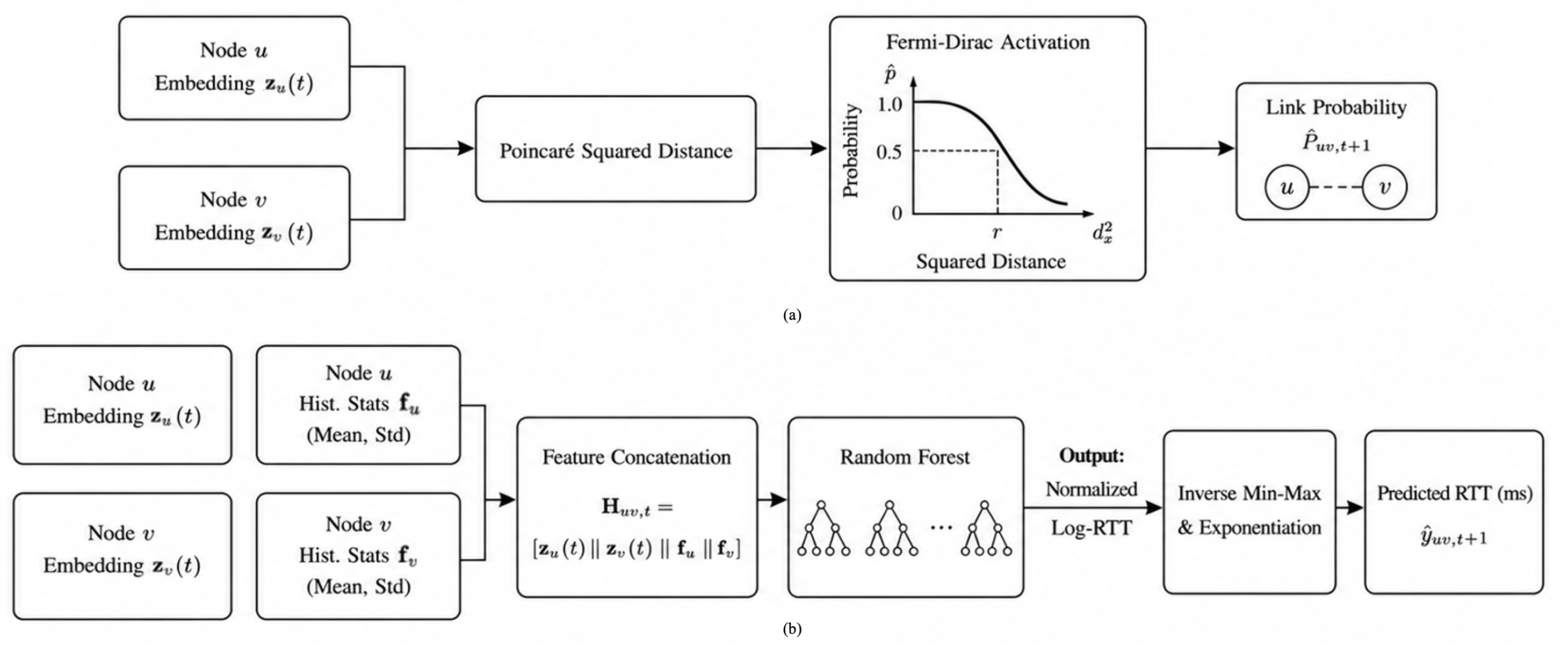}
    \caption{(a) Link prediction pipeline. Hyperbolic node embeddings are evaluated using the Poincaré squared distance, then mapped to link existence probabilities via Fermi-Dirac activation. (b) RTT prediction pipeline. Hyperbolic node embeddings and historical statistics are concatenated and processed by Random Forest to predict RTT values. Since the model is trained on normalized log-RTT targets, the output is subsequently denormalized via an inverse Min-Max transformation and exponentiated to recover the actual RTT in milliseconds.}
    \label{fig:link_rtt}
\end{figure*}

As illustrated in Fig.~\ref{fig:link_rtt}(a), for each candidate edge $(u,v)$ at timestamp $t$, we compute the Poincaré squared distance between node embeddings $ \mathbf z_u(t) $ and $ \mathbf z_v(t) $. The distance is then mapped to its link existence probability $\hat{p}_{uv,t+1}$ for the next timestamp through the Fermi-Dirac activation function, which converts hyperbolic distances into probabilities.

For its prediction of RTT (Fig.~\ref{fig:link_rtt}(b)), our model extracts the embeddings of its endpoints with node-level historical RTT statistics. The hyperbolic node embeddings $\mathbf{z}_u(t)$ and $\mathbf{z}_v(t)$ are concatenated with historical RTT statistics $\mathbf{f}_u$ and $\mathbf{f}_v$ (aggregated from past snapshots $G_1, \cdots G_t$) to form an edge-level feature vector:
\[
    \mathbf{H}_{uv,t}
    = [\,\mathbf{z}_u(t) \,\|\, \mathbf{z}_v(t) \,\|\, \mathbf{f}_u \,\|\, \mathbf{f}_v\,].
\]
The historical statistics in $\mathbf{f}_u$ and $\mathbf{f}_v$ are indeed the mean and standard deviation of log-transformed RTT values from all past links in which nodes $u$ and $v$ appear, either as \texttt{source}  or \texttt{target}, respectively.
A Random Forest regressor takes $\mathbf{H}_{uv,t}$ as input to predict log-RTT targets that are globally normalized over the training set.
Finally, the model performs an inverse Min-Max transformation and exponentiates them to recover the predicted RTT value $\hat{y}_{uv,t+1}$ in milliseconds. 

Our model jointly exploits the hyperbolic network topology and the temporal RTT dynamics.
\label{sec:vb3}
To jointly optimize topology learning and latency prediction, we train the hyperbolic encoder with a combined objective $\mathcal{L}$ that balances link prediction and RTT prediction:
\begin{equation}
    \mathcal{L} = \mathcal{L}_{\text{link}} + \lambda\,\mathcal{L}_{\text{RTT}},
    \label{eq:overall_loss}
\end{equation}
where $\mathcal{L}_{\text{link}}$ is the binary cross-entropy loss for link prediction, $\mathcal{L}_{\text{RTT}}$ is the MSE loss on the normalized RTT values, and $\lambda$ is a weighting hyperparameter. We set $\lambda = 10$ after exhaustively exploring different values to balance these two objectives. Here, our model minimizes this overall loss $\mathcal{L}$ using the Riemannian Adam (RAdam) optimizer~\cite{becigneul2023riemannian}.

The link prediction loss $\mathcal{L}_{\text{link}}$ is the binary 
cross-entropy between the predicted link probabilities $\hat{p}_{uv,t+1}$ (from 
the Fermi-Dirac decoder) and the ground-truth labels.
\begin{equation}
\begin{aligned}
    \mathcal{L}_{\text{link}} 
    &= -\frac{1}{|\mathcal{S}_{t+1}|} \sum_{(u,v) \in \mathcal{S}_{t+1}} 
    \Big[ \ell_{uv,t+1} \log \hat{p}_{uv,t+1} \\
    &\quad + (1-\ell_{uv,t+1}) \log(1-\hat{p}_{uv,t+1}) \Big]
\end{aligned}
\end{equation}
where $\mathcal{S}_{t+1}$ denotes the set of sampled node pairs (with some specific sampling method that will be mentioned later, $\mathcal{S}_{t+1}$ includes both 
positive edges that exist and negative samples that do not), and 
$\ell_{uv,t+1} \in \{0, 1\}$ is the binary label indicating whether edge $(u,v)$ exists at time $t+1$.

During the training phase, our model uses an auxiliary 2-layer MLP with hidden dimension 32 that takes the concatenated hyperbolic node embeddings $\mathbf{z}_{uv}(t) = [\mathbf{z}_u(t) \| \mathbf{z}_v(t)]$ at time $t$ as input to predict future RTT values at $t+1$, and optimizes the MSE loss $\mathcal{L}_{\text{RTT}}$, which is defined as follows
\begin{equation}
    \mathcal{L}_{\text{RTT}} 
    = \frac{1}{|E_{t+1}|}
      \sum_{(u,v)\in E_{t+1}} 
      \bigl( {y}_{uv,t+1} - \hat{y}^{\text{MLP}}_{uv,t+1} \bigr)^2
\end{equation}
where $E_{t+1}$ denotes the set of directed edges with RTT values that occurs at time $t+1$, $y_{uv, t+1} \in [0, 1]$ is the ground-truth RTT that in the normalized logarithmic form, and $\hat{y}^{\text{MLP}}_{uv,t+1} \in [0,1]$ is the MLP prediction passed through a sigmoid output layer, which is also in the same form.\shih{where is the MLP and sigmoid in it} This auxiliary task ensures that the node embeddings capture latency dynamics alongside network topology.


\label{training and inference details}

During training, we process the network snapshots sequentially to optimize the overall objective $\mathcal{L}$ defined in equation~\eqref{eq:overall_loss}. For each snapshot $G_t$, we use all  observed links $E_t$ as positive edges and construct an equal number of negative edges (1:1 ratio) by uniformly sampling unconnected node pairs from $V_t \times V_t \setminus E_t$. Our model then computes the link prediction loss $\mathcal{L}_{\text{link}} $ and the auxiliary RTT prediction loss $\mathcal{L}_{\text{RTT}}$ in parallel.

We split the dynamic graph into 85\% training, 5\% validation, and 10\% test snapshots in strict temporal order. To prevent data leakage, historical node-level RTT statistics are computed exclusively from the training set. During inference, we keep the model parameters fixed, but sequentially update the temporal encoder's hidden states over the validation and test snapshots to capture the latest network dynamics from the incoming snapshots. For link prediction, our results are evaluated using AUC and AP, whereas for RTT prediction, they are evaluated using RMSE and MAE on the original latency scale.


\label{robustness}
\label{fusion}

\section{Experiment}
\label{experiment}
All experiments are conducted on a Linux server equipped with an NVIDIA GeForce RTX 4090 GPU (24 GB) and an Intel Core i9-14900KF CPU. The models are implemented in Python 3.9 with PyTorch 2.x. We use a random seed of 1024 to ensure reproducibility.
We evaluate our method on the CAIDA IPv4 Ark Project dataset that spans 2015 to 2024. After the sampling in Section~\ref{dataset}, we obtain 1,456 daily snapshots, each representing the network routing and the corresponding RTT measurements on a specific date. To handle the heavy-tailed RTT distribution, we apply a logarithmic transformation followed by min-max normalization to $[0, 1]$. We adopt an 85\%/5\%/10\% data split, resulting in 1,241 snapshots for training, 70 for validation, and 145 for testing, ensuring temporal integrity and preventing information leakage.

We compare HERMIT against baselines in two categories. For link prediction, we evaluate against HTGN~\cite{yang2021discrete}, which integrates hyperbolic graph convolution with gated recurrent units, and HMPTGN~\cite{le2024toward}, our base encoder that performs manifold-preserving message passing but lacks the edge encoder and hybrid RTT prediction components proposed in this work. For RTT prediction, we compare against Random Forest trained on tabular features only~\cite{stepanov2025round}, whereas our HERMIT enhances the Random Forest regressor with additional hyperbolic embeddings that encode network topology.

The hyperbolic embedding of our HERMIT model is implemented using a Poincaré ball manifold with curvature $c=1.0$. Edge features are 3-dimensional vectors that consist of the log-RTT mean, the log-RTT standard deviation, and the link connection count (link weight). The edge encoder then applies a linear transformation on these edge features from three dimensions to one with sigmoid activation. During training, the auxiliary head is a two-layer MLP with 32 hidden units\shih{for shih: MLP details great}, ReLU activation, a dropout rate of 0.2, and a sigmoid output layer, which is used to calculate $\mathcal{L}_{\text{RTT}}$. We use a learning rate of 0.0001 and set the RTT weight in loss ($\lambda$) to 10.0. For the final RTT prediction, the Random Forest component uses 120 decision trees with maximum depth 30 and maximum features 0.8, optimized via grid search on the validation set, and it takes concatenated hyperbolic embeddings (that are projected to tangent space) and tabular features (historical RTT statistics) as input.

Models are trained for up to 50 epochs with early stopping based on validation performance. We evaluate link prediction using Area Under the ROC Curve (AUC), which measures how well the model separates existing from non-existent links, and New AUC, computed only on edges appearing for the first time in test snapshots to assess generalization to unseen links. For RTT prediction, we report Root Mean Squared Error  (RMSE) and Mean Absolute Error (MAE). \shih{full name mentioned first time, then later just write the abbreviated name. apply for all terms.} RMSE penalizes large errors heavily and is sensitive to outliers, while MAE reflects typical accuracy under normal conditions. Together, they capture both average performance and robustness under high-variance network conditions.

\section{Results}
\label{results}
We evaluate the link prediction performance of HERMIT against two baselines: HTGN and HMPTGN. Table~\ref{tab:lp} shows that our model achieves an AUC of 99.53\% and AP of 99.56\% on the test set, outperforming HTGN (96.28\% AUC, 96.47\% AP) and HMPTGN (97.13\% AUC, 97.58\% AP). For new link prediction, HERMIT attains a new AUC of 99.32\% and new AP of 99.38\% (Table~\ref{tab:lp}), demonstrating strong generalization to previously unseen connections. The performance improvement can be attributed to the incorporation of edge-level features, which enable the model to capture fine-grained temporal dynamics and link quality information. Table~\ref{tab:rtt_prediction_results} compares the RTT prediction performance between HERMIT and Random Forest trained only on tabular features (historical RTT statistics). On the complete test set, HERMIT achieves an MAE of 5.35 ms and RMSE of 20.85 ms, representing improvements of 1.3\% and 6.4\% over the baseline, respectively. The larger RMSE improvement indicates the model is particularly effective at reducing large prediction errors. For new links that are unseen during training, HERMIT maintains its advantage with MAE of 6.85 ms compared to 6.94 ms (1.3\% improvement), demonstrating that hyperbolic embeddings provide generalizable structural information for previously unseen connections.
\begin{table}[h!]
    \centering
    \caption{Link prediction and new link prediction performance on CAIDA dataset}
    \label{tab:lp}
    \resizebox{0.5\textwidth}{!}{%
    \begin{tabular}{lccc}
        \toprule
        \textbf{Metric} & \textbf{HTGN} & \textbf{HMPTGN} & \textbf{HERMIT} \\
        \midrule
        AUC & 0.9628 & 0.9713 & \textbf{0.9953} \\
        AP & 0.9647 & 0.9758 & \textbf{0.9956} \\
        New AUC & 0.9517 & 0.9654 & \textbf{0.9932} \\
        New AP & 0.9548 & 0.9705 & \textbf{0.9938} \\
        \bottomrule
    \end{tabular}
    }
\end{table}

\begin{table}[h!]
    \centering
    \caption{RTT Prediction Performance Comparison (Test Set)}
    \label{tab:rtt_prediction_results}
    \resizebox{0.5\textwidth}{!}{%
    \begin{tabular}{l l c c}
        \toprule
        \textbf{Category} & \textbf{Model} & \textbf{MAE (ms)} & \textbf{RMSE (ms)} \\
        \midrule
        \multirow{2}{*}{Global} 
        & HERMIT& \textbf{5.35}& \textbf{20.85} \\
        & Random Forest& 5.42 & 22.18 \\
        \midrule
        \multirow{2}{*}{Existing} 
        & HERMIT& \textbf{5.28}& \textbf{20.73}\\
        & Random Forest& 5.35 & 22.05 \\
        \midrule
        \multirow{2}{*}{New} 
        & HERMIT& \textbf{6.85}& \textbf{24.52} \\
        & Random Forest& 6.94 & 25.09 \\
        \bottomrule
    \end{tabular}
    }
    \vspace{2pt}
    {\small
    \textit{Note.} Global: all test edges; Existing: links observed during training; 
    New: previously unseen links. 
    }
\end{table}

Table~\ref{tab:convergence} shows the training convergence behavior across 50 epochs. The model exhibits rapid convergence within the first 10 epochs, achieving an AUC above 0.99. Performance stabilizes after epoch 20, with the best model obtained at epoch 30 (AUC = 0.9953, New AUC = 0.9932). Performance slightly decreases at epochs 40 and 50, suggesting that early stopping at epoch 30 is appropriate to prevent overfitting while maintaining strong generalization. The consistent gap between overall AUC and New AUC (approximately 0.2\%) throughout training indicates stable generalization capability across all epochs.

\begin{table}[t]
\centering
\begin{threeparttable}
    \caption{HMPTGN training convergence (Learning Rate = 0.0001)}
    \label{tab:convergence}
    \setlength{\tabcolsep}{20pt} 
    \renewcommand{\arraystretch}{1}
    \begin{tabular}{ccc}
        \toprule
        \textbf{Epoch} & \textbf{Test AUC} & \textbf{New AUC} \\
        \midrule
        1  & 0.9520 & 0.9471 \\
        5  & 0.9936 & 0.9918 \\
        10 & 0.9963 & 0.9946 \\
        20 & 0.9946 & 0.9919 \\
        \textbf{30} & \textbf{0.9953} & \textbf{0.9932} \\
        40 & 0.9945 & 0.9919 \\
        50 & 0.9943 & 0.9918 \\
        \bottomrule
    \end{tabular}

    \begin{tablenotes}
        \footnotesize
        \item Note: The best model is selected at epoch 30 based on validation loss.
    \end{tablenotes}
\end{threeparttable}
\end{table}

\section{Discussion}
\label{discussion}
Our results highlight the value of the hybrid HERMIT architecture, which combines hyperbolic temporal graph representations with a Random Forest regressor. This design leverages both topological structure and statistical RTT features, achieving more accurate predictions than either approach alone. The hybrid model demonstrates strong generalization performance on unseen edges, indicating that the hyperbolic embeddings capture latent hierarchical structures that effectively infer unseen network connections. Incorporating edge-level features, such as RTT statistics (mean and standard deviation) and link weights, consistently improves prediction accuracy by providing complementary information that node embeddings alone cannot capture. Furthermore, manifold-preserving message passing effectively models long-range routing paths, since operating directly in hyperbolic space helps preserve the geometric characteristics of the evolving Internet topology.

From a deployment perspective, HERMIT offers significant practical advantages. The model achieves scalability by operating on 16-dimensional hyperbolic embeddings, which efficiently encode hierarchical routing structures in networks with thousands of nodes and edges.  While hyperbolic operations introduce computational overhead, this cost occurs only during offline training on historical data. Once trained, the model efficiently deploys through a two-stage pipeline: First, the hyperbolic temporal encoder processes network snapshots to generate hyperbolic embeddings. Second, the Random Forest regressor combines these embeddings with RTT statistics to produce predictions. For static or slowly evolving topologies, embeddings can be pre-computed and cached, allowing the Random Forest to perform fast cached inference without re-encoding. This design is particularly suited to operational scenarios such as latency-aware traffic engineering and proactive performance monitoring that require frequent, low-latency RTT predictions.


\section{Conclusion}
\label{conclusion}
We propose HERMIT, a hybrid framework combining hyperbolic temporal graph neural networks with Random Forest regression for joint link prediction and RTT prediction on dynamic Internet topologies. By incorporating edge-level RTT features and a learnable edge encoder into the hyperbolic temporal encoder, and fusing the resulting hyperbolic embeddings with historical RTT statistics, our model achieves superior link prediction performance and stronger and more robust RTT prediction accuracy on both existing and newly emerging links in a ten-year CAIDA dataset. For future work, we will extend the framework to multi-metric network performance prediction and validate its robustness across diverse network environments and online deployment scenarios.

\newpage
\bibliographystyle{IEEEtran}
\bibliography{reference}











\end{document}